\documentclass[11pt]{article}

\usepackage[preprint]{acl}

\usepackage{times}
\usepackage{latexsym}
\usepackage{amsmath}
\usepackage{amsfonts}

\usepackage[T1]{fontenc}

\usepackage[utf8]{inputenc}

\usepackage{microtype}

\usepackage{graphicx}
\usepackage{booktabs}
\usepackage{placeins}
\usepackage{algorithm}
\usepackage{algorithmic}
\usepackage{xspace}
\usepackage{enumitem}
\usepackage{listings}
\usepackage{threeparttable}

\newcommand{\ours}{\textsc{HoliDivEvolve}\xspace}

\title{Holistic Diversity Management in LLM-Guided Evolutionary Search}

\author{
\parbox{\dimexpr\textwidth-2\tabcolsep\relax}{\centering
\!\!Seockbean Song$^{1}$,\!  Youngsik Yoon$^{2}$,\! Sungjae Lee$^{2}$,\! Siwei Wang$^{3}$,\! Wei Chen$^{3}$,\! Jungseul Ok$^{1,2}$\thanks{Correspondence to: jungseul.ok@postech.ac.kr} \\ [0.3em]
\mdseries
$^{1}$Graduate School of AI, POSTECH,
$^{2}$Department of CSE, POSTECH, \\
$^{3}$Microsoft Research Asia \\ [0.3em]
\text{\{shinebobo, ysyoon97, sungjaelee25, jungseul.ok\}@postech.ac.kr}, \\[0.3em] 
\text{\{siweiwang, weic\}@microsoft.com}
}}

\begin{document}
\maketitle

\begin{abstract}
\label{sec:abstract}
LLM-guided evolutionary search iteratively selects existing programs, uses LLMs as mutation operators to generate a new program, and evaluates it to guide subsequent search.
To preserve diversity and reduce the risk of premature convergence, it often adopts the \textit{island model}, in which groups of programs, called islands, evolve separately.
However, this structure alone does not guarantee distinct search directions, as islands may start from similar programs or converge toward similar solutions during evolution.
This suggests that diversity must be considered throughout the search process, from initializing islands with distinct programs to developing island trajectories and sharing discoveries across islands.
To this end, we propose \ours, a holistic approach to establishing and effectively utilizing island diversity throughout evolutionary search.
Specifically, we introduce diverse initialization to establish different initial programs, trajectory adaptive island selection to allocate computation based on current score, recent growth, and the remaining search budget, and growth adaptive migration to share alternative ideas across islands.
Experiments across diverse benchmarks show that \ours achieves competitive performance on most evaluated tasks.
We further conduct controlled analyses to examine how our design choices affect search behavior and final performance.
\end{abstract}

\section{Introduction}
\label{sec:intro}

Large language models (LLMs) are increasingly used as the mutation operator inside evolutionary search, turning program and algorithm discovery into an iterative loop of generation, evaluation, and selection \citep{romera2024mathematical,novikov2025alphaevolve}. 
To mitigate premature convergence, the island model is commonly adopted as a classical approach to maintaining diversity \citep{tanese1989distributed}.
This model organizes candidate programs into groups called \textit{islands}, which evolve separately and occasionally
exchange programs through migration.

However, the island model does not directly control the diversity of programs generated within each island. 
This problem gets worse when LLMs are used as mutation operators, because LLMs often generate programs with similar algorithms
\citep{brownlee2025large, lee2025diversely, gurkan2026mutation}.

Prior studies have addressed diversity through different aspects of the island model. 
Some approaches focus on diverse initialization or population management \cite{yang2026turboevolve,tanveer2026levi, lange2026shinkaevolve}, while others adapt island selection \cite{cemri2026adaevolve,luo2026relay} or migration \cite{yan2026pacevolve,li2026topofe}. 
However, these decisions have often been studied separately, despite their interconnected roles in maintaining diversity throughout search.

We argue that establishing and utilizing island diversity should be considered together. 
Distinct initial programs provide different search directions, but their subsequent development depends on how computation is allocated across islands. 
Likewise, migration allows islands to benefit from discoveries made elsewhere, while also influencing their future search directions. 
These decisions therefore shape one another throughout evolution, motivating a holistic approach to island diversity.

Based on this view, we propose \ours, which takes a holistic approach to establishing and effectively utilizing island diversity throughout evolutionary search.
We generate multiple candidate programs and use a diverse subset to initialize different islands, establishing distinct starting points.
We then allocate computation according to each island's current score, recent growth, and the remaining budget, allowing different island trajectories to develop.
We further guide cross-island migration using relative growth, introducing alternative programs when an island's growth lags behind others.
In this way, \ours considers not only how diverse island trajectories are established, but also how they develop and benefit from discoveries made elsewhere.

We evaluate \ours across a diverse set of program and algorithm optimization tasks and compare it with recent LLM-guided evolutionary methods. Overall, \ours achieves competitive performance and outperforms existing approaches on most evaluated tasks. Our component studies further examine the effects of each component, showing how it influences the resulting search process and final solution quality.

\begin{figure*}[!t]
    \centering
    \includegraphics[width=\textwidth]{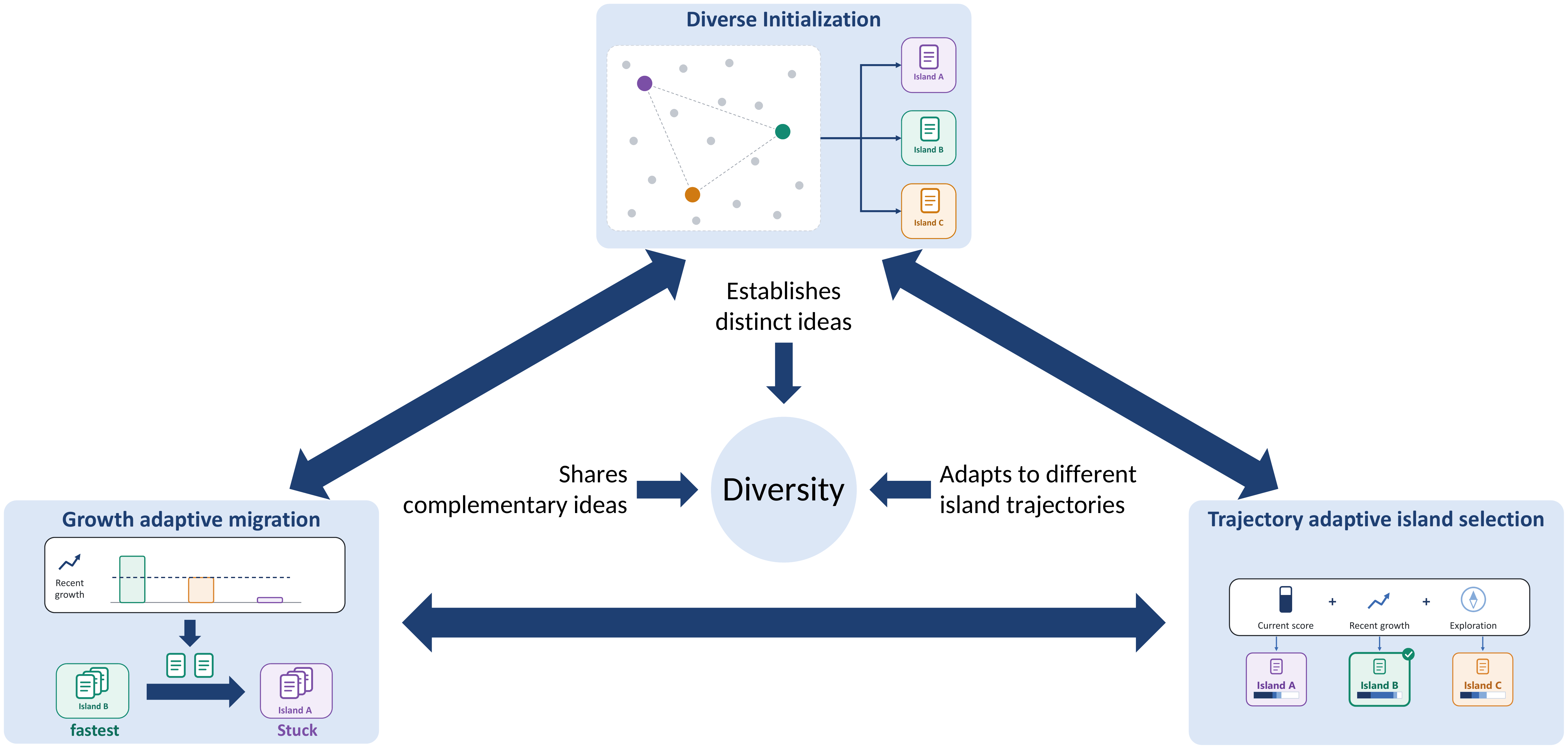}
    \caption{Overview of \ours. We consider three components for diversity management: diverse initialization, trajectory adaptive island selection, and growth adaptive migration.}
    \label{fig:algorithm_overview}
\end{figure*}

We summarize our contributions as follows:
\begin{itemize}
    \item We propose a \textbf{diverse initialization} strategy that establishes distinct search directions by assigning
    different initial programs to different islands.

    \item We develop a \textbf{trajectory adaptive island selection} strategy that allocates computation based on current
    score, recent growth, and the remaining search budget to effectively develop diverse island trajectories.

    \item We introduce a \textbf{growth adaptive migration} strategy that enables islands to share alternative ideas while maintaining their existing search directions.
\end{itemize}
\section{Related Work}
\label{sec:related_work}

\paragraph{LLM-guided evolutionary search.}
LLMs have increasingly been used as mutation operators for program and
algorithm discovery, where generated solutions are evaluated and reused
to guide subsequent generations
\citep{romera2024mathematical,liu2024evolution, ye2024reevo, novikov2025alphaevolve,openevolve, jiang2026deltaevolve, assumpccao2025codeevolve, tanveer2026levi,liu2602evox}.
These works have extended evolutionary search through improved population
management, reflection, search history, and mutation strategies.
Our work builds on this paradigm but focuses specifically on how multiple
island trajectories should be initialized, developed, and connected
throughout a finite search budget.

\paragraph{Adaptive search allocation.}
Recent methods make different parts of evolutionary search adaptive using
feedback collected during evolution, including model selection, parent
sampling, search-strategy selection, and computation allocation
\citep{lange2026shinkaevolve,cemri2026adaevolve,chen2026tide,luo2026relay}.
Our setting instead considers computation allocation across islands whose
search trajectories may improve at different rates.
Our island selection rule is motivated by best arm identification for rising bandits~\citep{metelli2022stochastic,song2026rising,mussi2023best}. In this setting, finding the option that will be best in the future requires considering its current score, its recent growth, and the remaining budget.

\paragraph{Diversity and cross-island interaction.}
Maintaining diverse subpopulations and exchanging information between
them are central ideas in island-based evolutionary algorithms \cite{tanese1989distributed}.
Recent LLM-guided methods further explore diversified initialization,
parallel trajectories, and adaptive information sharing \citep{yang2026turboevolve,yan2026pacevolve,li2026topofe}.
Our method initializes islands from separated programs and subsequently
uses relative recent growth to determine when an island should receive
programs from another trajectory.

\section{Problem Formulation}
\label{sec:problem}

We consider the problem of discovering high-scoring programs using an LLM under a total budget of $T$ program generations, including those used for initialization.
Let $\mathcal{X}$ denote the space of candidate programs and
$f:\mathcal{X}\to\mathbb{R}$ be a fixed evaluation function,
where higher scores indicate better performance.
The search maintains $K$ islands, indexed by
$[K] := \{1,\ldots,K\}$.
Each island $k$ maintains a population $\mathcal{D}_{k,t}$
containing program and score pairs available at the end of round $t$.
The process starts from a single evaluated initial program
$(X_0,s_0)$, where $s_0=f(X_0)$, shared across all islands:
$\mathcal{D}_{k,0}=\{(X_0,s_0)\}$ for all $k\in[K]$.

At each round $t$, we select an island $I_t\in[K]$.
A stochastic prompt builder $PB$ then constructs a prompt
from the selected island's population:
\begin{align*}
    P_t \sim PB(\mathcal{D}_{I_t,t-1}).
\end{align*}
Here, $PB(\mathcal{D})$ denotes a distribution over prompts
constructed from the programs and evaluation scores stored in
$\mathcal{D}$.
The prompt-building procedure may include selecting parent or
reference programs and incorporating their scores into the prompt.
We follow the prompt construction procedure of \citet{lange2026shinkaevolve, cemri2026adaevolve} to construct mutation prompts from the selected island's population.

Conditioned on the sampled prompt $P_t$, the LLM generates
an offspring program:
\begin{align*}
    X_t \sim LLM(P_t),
\end{align*}
where $LLM(P)$ denotes the distribution over candidate programs
induced by prompting the LLM with $P$.
Thus, the generation process accounts for randomness in both
prompt construction and LLM generation.

We evaluate the offspring to obtain score $s_t=f(X_t)$ and add
the resulting program and score pair to the selected island.
The populations are updated as $\mathcal{D}_{I_t,t} = \mathcal{D}_{I_t,t-1}\cup\{(X_t,s_t)\}$.
Consequently, each island retains its initial records and
all offspring evaluated during rounds in which it was selected.
These records provide the information from which prompts
can be constructed in subsequent rounds.
Migration allows islands to use programs discovered on other islands to guide subsequent program generation.
We incorporate this mechanism by allowing evaluated program records to be transferred between island pools.
After evaluating $X_t$ and updating the selected island, let
$\widetilde{\mathcal D}_{1:K,t}$ denote the collection of island
pools before migration. A migration policy $\mathcal M_t$ may then
exchange previously evaluated programs across islands:
\begin{align*}
    \mathcal D_{1:K,t}
    =
    \mathcal M_t\!\left(
        \widetilde{\mathcal D}_{1:K,t}
    \right).
\end{align*}
Here, $\mathcal M_t$ determines whether migration occurs, which
islands exchange programs, and which programs are incorporated into
the recipient pools. If no migration occurs, $\mathcal M_t$ leaves
the pools unchanged. Migration does not itself generate a new program
or consume a search round; instead, it changes the programs available
for subsequent prompt construction.

Our objective is to maximize the expected score of the best
program discovered within the search budget, rather than
the average or cumulative score of all generated programs.
For each round $t$, define the best score across all islands as follows:
\begin{align*}
    B_t := \max_{l\leq t} s_l  \;.
\end{align*}
At the end of the search, we return a program $\widehat{X}_T$
attaining $B_T$, with ties broken arbitrarily.
Let $\mathcal{A}$ denote a search procedure implementing
the above process.
We measure its performance by
\begin{align*}
    J_T(\mathcal{A})
    :=
    \mathbb{E}_{\mathcal{A}}
    \left[B_T\right]
    .
\end{align*}
The expectation is taken over the randomness of the search process induced
by stochastic prompt construction, stochastic LLM generation,
and any randomness in island selection.
Our goal is to design a search procedure that maximizes
$J_T(\mathcal{A})$ subject to the specified information
and budget constraints.

Thus, each search run is evaluated by its single best
discovered program, and performance is averaged over
the randomness of repeated runs.
The returned program need not be the last generated program
and may originate from any island or from the initial populations.
\section{Method}
\label{sec:method}

\begin{algorithm}[t]
\caption{\ours}
\label{alg:ours}
\begin{algorithmic}
\REQUIRE LLM, prompt builder $PB$, evaluator $f$, number of islands $K$, number of initial candidates $N$, total budget $T$

\item[\textbf{Initialization:}]
\STATE Generate and evaluate $N$ candidate programs
\STATE Select $K$ initial programs based on equation~\ref{eq:diverse_initialization} and initialize the $K$ islands

\FOR{$t=1,\ldots,T-N$}
    \STATE Select island $I_t$ based on equation~\ref{eq:island_selection}
    \STATE Sample $P_t \sim PB(\mathcal D_{I_t,t-1})$
    \STATE Generate $X_t \sim LLM(P_t)$ and evaluate $s_t=f(X_t)$
    \STATE Update island $I_t$ and its growth statistics

    \IF{migration is triggered (Section~\ref{sec:migration})}
        \STATE Select and migrate programs according to Section~\ref{sec:migration}
    \ENDIF
\ENDFOR

\RETURN the best program found
\end{algorithmic}
\end{algorithm}

We organize our method around three decisions: how to initialize the
islands, which island to select at each round, and when and what to
exchange across islands.
We address these decisions with three components:
\textit{diverse initialization}, which provides different initial
programs; \textit{trajectory adaptive island selection}, which allocates
search rounds based on the state of each island; and
\textit{growth adaptive migration}, which adaptively migrates programs across islands when an island's recent growth slows relative to others.
Figure~\ref{fig:algorithm_overview} summarizes the overall procedure.

\subsection{Diverse Initialization}
\label{sec:initialization}

We allocate the first $N$ program generations to producing $N$ candidate programs using prompts that encourage different search directions, before beginning evolutionary search.
We evaluate the candidates and maintain the valid programs in
a set $\mathcal V$. 
Each program is represented using a code embedding model, and cosine distance is used to measure the distance between programs.

From $\mathcal V$, we select $K$ programs that are well separated from
one another:
\begin{align}
    S^\star
    \in
    \arg\max_{\substack{S\subseteq\mathcal V\\|S|=K}}
    \;
    \min_{\substack{p,q\in S\\p\neq q}}
    d(p,q) \;,
    \label{eq:diverse_initialization}
\end{align}
where $d(p,q)$ denotes the cosine distance between programs $p$ and $q$.

For a candidate set $S$, the inner minimum measures the distance between
the most similar pair of programs in the set. The outer maximization
selects the set for which this minimum distance is largest.
Therefore, this criterion encourages each island to start from a distinct program and 
explore a different island trajectory.
We assign one selected program to each
island, providing the islands with distinct starting points before the
evolutionary search begins.

\subsection{Trajectory Adaptive Island Selection}
\label{sec:selection}

Diverse initialization encourages islands to follow different island
trajectories. 
Since these trajectories can differ in both their current
quality and recent improvement, the search should allocate computation
according to how each island develops rather than treating all islands
uniformly. 
Our key intuition is to balance current score and recent growth
according to the remaining search budget, emphasizing growth when there
is more time for further improvement and current quality as the budget
is exhausted. 
We define the evolutionary search horizon as  $H=T-N$ and $w_t=(H-t)/H$ and select the next island
at round $t$ according to
\begin{align}
    I_{t}\in\arg\max_{i\in[K]}
    \Big[
    &\underbrace{(1-w_{t-1})\,\widetilde{\mu}_i(t-1)}_{\text{current score}}  \notag \\
    +&\underbrace{w_{t-1}\,\widetilde{g}_i(t-1)}_{\text{recent growth}}  \notag \\
    +&\underbrace{
        C \sqrt{\frac{\log(t+1)}{n_i(t-1)}}
    }_{\text{exploration bonus}}
    \Big] \;, 
    \label{eq:island_selection}
\end{align}
where, $n_i(t)$ counts evaluations assigned to island $i$
up to round $t$, and $C$ controls exploration of
less-visited islands.

\paragraph{Current score.}
We estimate the current score $s$ of island $i$ by averaging the scores of programs in its current population:
\begin{align*}
    \hat{\mu}_i(t) := \frac{1}{|\mathcal D_{i,t}|} \sum_{(p,s)\in\mathcal D_{i,t}} s \;.
\end{align*}
We use the population average because future prompts can draw both
parent and reference programs from the island, so the overall quality
of its current programs affects subsequent mutations.
We normalize $\hat{\mu}_i(t)$ across islands and denote the resulting
value by $\widetilde{\mu}_i(t)$.

\paragraph{Recent growth.}
Recent improvement is widely used to summarize search progress in
adaptive evolutionary search
\citep{cemri2026adaevolve, yan2026pacevolve}.
We similarly use the improvements observed from recent mutations to
estimate the recent growth of each island trajectory.

When island $I_t$ is selected at round $t$, let $s^*_{I_t}(t-1)$ denote the best score 
available in island $I_t$ before round $t$. We define its normalized improvement as follows:
\begin{align*}
    \delta_{I_t}(t) =\min\left\{1,\,\frac{\max\{0,s_t-s^*_{I_t}(t-1)\}} {|s^*_{I_t}(t-1)|+\epsilon}\right\},
\end{align*}
where $\epsilon>0$ provides numerical stability.

We update the recent growth estimate of the selected island using an exponentially weighted average:
\begin{align*}
    \hat{g}_{I_t}(t)^2 = \rho\,\hat{g}_{I_t}(t-1)^2
        +(1-\rho)\delta_{I_t}(t)^2,
\end{align*}
where $\rho\in(0,1)$ determines how quickly older observations are
discounted.
For islands not selected at round $t$, the estimate remains unchanged.

A larger $\hat{g}_i(t)$ therefore indicates that island $i$ has
produced larger improvements in its recent evaluations, while the
estimate decreases when repeated mutations fail to improve its best
program. We normalize $\hat{g}_i(t)$ across islands and denote the
normalized value by $\widetilde{g}_i(t)$.
We additionally adopt an LLM-based exploration mechanism from prior work \citep{cemri2026adaevolve}, which generates new ideas when the search remains stagnant for an extended period.

\subsection{Growth Adaptive Migration}
\label{sec:migration}

Island selection determines which island receives additional computation, but without migration, islands evolve largely as independent refinement processes. Migration allows programs discovered in one island to influence subsequent mutations in others. We employ two forms of migration: conventional periodic migration for occasional program sharing across islands, and growth adaptive migration to introduce alternative programs when an island's recent growth becomes relatively low. We focus on growth adaptive migration below, with implementation details of periodic migration provided in Appendix~\ref{app:migration}.

\paragraph{When and where to migrate.}
After island $I_t$ is updated at round $t$,
we compare its recent growth with that of other islands and trigger migration when

Specifically, we trigger migration when
\begin{align*}
    \hat{g}_{I_t}(t)
    \le
    \tau \max_{j\in[K]} \hat{g}_j(t),
\end{align*}
where $\tau\in(0,1)$ controls how much slower the recipient must be
growing before migration is triggered.

When migration occurs, we choose the donor as
\begin{align*}
    j^*
    \in
    \arg\max_{j\neq I_t}
    \hat{g}_j(t).
\end{align*}
Thus, an island whose recent growth has slowed receives programs from
an island that is currently exhibiting stronger improvement.

\paragraph{Which programs to migrate.}
Once a donor island $j^*$ is selected, we do not simply transfer its
highest-scoring programs. 
Doing so would mainly propagate the donor's current solution and can make the recipient follow a similar search
direction. 
Instead, we use migration to introduce alternative ideas while
preserving the recipient's island trajectory.

Let $p^*_{I_t}(t)$ denote the current best program in the recipient
island. Among the eligible non-best programs in the donor, we prioritize
programs that are most different from $p^*_{I_t}(t)$, measured by cosine
distance in the code-embedding space. 
We then transfer the highest-ranked programs to the recipient. 
In this way, the migrated programs are chosen relative to what the recipient has already discovered, rather than by
their score in the donor island alone.

\paragraph{How migrated programs are used.}
Migrated programs are used to complement the recipient's current search
rather than replace it.
When the recipient island is selected again, its current best program
remains the parent, while the migrated programs are prioritized as
references in the prompt.
This separation between parent and reference preserves the recipient's
current direction while allowing external ideas to influence its
subsequent mutations.

\paragraph{Updating growth after migration.}
Migration changes the programs available in the recipient island, so its
pre-migration growth estimate may no longer fully represent its updated
search state. At the same time, migration itself should not be counted as
an improvement produced by a local mutation. We therefore keep the visit
count unchanged and update the recipient's growth estimate according to
the fraction of programs introduced from the donor.

Let $q$ denote the number of migrated programs and $P_{I_t}$ the
recipient population size before migration. We update
\begin{align*}
    \hat{g}_{I_t}(t)^2
    \gets
    \left(1-\frac{q}{P_{I_t}}\right)\hat{g}_{I_t}(t)^2
    +
    \frac{q}{P_{I_t}}\hat{g}_{j^\star}(t)^2.
\end{align*}
Thus, larger transfers place more weight on the donor's recent growth,
while smaller transfers preserve more of the recipient's previous growth
estimate.

\begin{table*}[!t]
\begin{threeparttable}
\centering
\footnotesize
\setlength{\tabcolsep}{2.5pt}
\renewcommand{\arraystretch}{1.15}

\begin{tabular*}{\textwidth}{@{\extracolsep{\fill}}llrrrrrr@{}}
\toprule
& &
\multicolumn{3}{c}{Mathematical Optimization} &
\multicolumn{3}{c}{Systems Optimization} \\
\cmidrule(lr){3-5}
\cmidrule(lr){6-8}

Backbone & Algorithm &
\shortstack{Heilbronn\\Triangle $\uparrow$} &
\shortstack{Circle Packing\\(Square) $\uparrow$} &
\shortstack{Signal\\Processing $\uparrow$} &
PRISM $\uparrow$ &
EPLB $\uparrow$ &
CloudCast $\downarrow$ \\
\midrule

GPT-5 & OpenEvolve
& $0.028\pm.006$ & $2.531\pm.018$
& $0.569\pm.047$ & $\underline{26.23}\pm.00$
& $0.127\pm.000$ & $926.9\pm171.0$ \\

& ShinkaEvolve
& $0.032\pm.012$ & $2.464\pm.083$
& $0.485\pm.044$ & $\mathbf{26.26}\pm.00$
& $0.118\pm.010$ & $954.8\pm125.0$ \\

& AdaEvolve
& $\underline{0.033}\pm.001$
& $\mathbf{2.629}\pm.001$
& $\underline{0.698}\pm.017$
& $\leq 26.26$\tnote{\dag}
& $\underline{0.134}\pm.010$
& $\underline{662.3}\pm.1$ \\

& \textbf{Ours}
& $\mathbf{0.034}\pm.002$
& $\underline{2.622}\pm.008$
& $\mathbf{0.722}\pm.013$
& $\mathbf{26.26}\pm.00$
& $\mathbf{0.146}\pm.000$
& $\mathbf{643.9}\pm36.6$ \\

\midrule

Qwen3-32B & OpenEvolve
& $0.012\pm.011$ & $1.970\pm.073$
& $0.531\pm.029$ & $23.55\pm.63$
& $0.127\pm.000$ & $1020.2\pm25.9$ \\

& ShinkaEvolve
& $\underline{0.019}\pm.008$
& $\underline{2.152}\pm.256$
& $0.560\pm.043$
& $\underline{25.40}\pm.28$
& $\underline{0.128}\pm.001$
& $1005.2\pm25.9$ \\

& AdaEvolve
& $\mathbf{0.022}\pm.002$
& $\mathbf{2.621}\pm.003$
& $\underline{0.622}\pm.076$
& $25.15\pm.97$
& $0.127\pm.001$
& $\mathbf{716.2}\pm20.1$ \\

& \textbf{Ours}
& $\mathbf{0.022}\pm.002$
& $\mathbf{2.621}\pm.001$
& $\mathbf{0.659}\pm.051$
& $\mathbf{25.84}\pm.20$
& $\mathbf{0.145}\pm.000$
& $\underline{751.2}\pm125.9$ \\

\bottomrule
\end{tabular*}

\caption{
Performance comparison across mathematical and systems
optimization benchmarks using GPT-5 and Qwen3-32B
(mean $\pm$ SD).
Bold indicates the best observed mean within each backbone,
and underline indicates the second-best distinct mean,
with ties determined at displayed precision.
GPT-5 baselines are from Tables 1--2 of
\citet{cemri2026adaevolve} (three runs).
All methods use a total budget of 100 program generations:
Ours allocates 20 generations to diverse initialization
and 80 rounds to evolutionary search, while the baselines
use all 100 rounds for search with the same initial program.
Arrows indicate the preferred direction.
}
\label{tab:main_gpt5}

\begin{tablenotes}[flushleft]
\footnotesize
\item[$\dagger$]
The original result was affected by an evaluator error.
We report the maximum possible score of 26.26 as an
upper bound, not as an observed mean.
\end{tablenotes}

\end{threeparttable}
\end{table*}

\section{Experiments}
\label{sec:experiments}
We first compare the overall performance of our method with recent
LLM-guided evolutionary approaches across multiple program optimization
tasks. 
We then conduct controlled studies to examine how our design choices affect search behavior and final performance.

\subsection{Experimental Setup}
\label{sec:experimental_setup}

We evaluate \ours on six program optimization benchmarks:
Heilbronn Triangle, Circle Packing, Signal Processing, PRISM,
EPLB, and CloudCast. Each benchmark uses a task-specific evaluator,
with higher scores indicating better performance except for
CloudCast, which reports cost. Detailed benchmark descriptions
and evaluation metrics are provided in Appendix~\ref{app:benchmarks}.

For overall performance comparisons, we report results using
GPT-5 and Qwen3-32B separately. We use Qwen3.8-27B for controlled
component studies on Signal Processing, examining diverse
initialization, island selection, and ablation.
We report the final best score as mean $\pm$ sample standard
deviation across independent search runs, except where otherwise
specified. Further experimental settings are provided in
Appendix~\ref{app:configuration}.

\subsection{Overall Performance}
\label{sec:overall_performance}

Table~\ref{tab:main_gpt5} compares our method with AdaEvolve,
OpenEvolve, and ShinkaEvolve in separate backbone blocks.
The GPT-5 baseline entries are drawn from
Tables 1--2 of \citet{cemri2026adaevolve}, which report mean
$\pm$ standard deviation over three runs with 100 search iterations.
We retain the published objective units.
For GPT-5, we evaluate \ours over three random seeds, allocating 20 program generations to initialization and 80 rounds to evolutionary search for each run.
For Qwen3-32B, we similarly report results over three random seeds for each benchmark.

\subsection{Diverse Initialization}
\label{sec:initialization_analysis}

\paragraph{Reducing cross-island redundancy.}
We first evaluate whether diverse initialization produces more diverse island trajectories than same-start initialization.
Table~\ref{tab:initialization_redundancy} compares same-start and max--min initialization on Signal Processing by measuring how similar the best programs and populations are across islands. 
We compute these cross-island diversity metrics after 20 evolutionary
rounds, before any migration occurs, so that the comparison reflects the effect of initialization rather than later information exchange.

\begin{table}[!htb]
\centering
\small
\setlength{\tabcolsep}{4pt}
\renewcommand{\arraystretch}{1.12}
\begin{tabular}{@{}p{0.55\linewidth}rr@{}}
\toprule
Metric $\downarrow$ & Same & Ours (Max--min) \\
\midrule
Best program similarity
    & 0.183 & \textbf{0.062} \\
Population proximity
    & 0.424 & \textbf{0.086} \\
Exact code overlap
    & 0.183 & \textbf{0.000} \\
Descendant proximity
    & 0.122 & \textbf{0.085} \\
\bottomrule
\end{tabular}
\caption{
Cross-island redundancy after 20 evolutionary rounds in Signal Processing under
same start and max--min initialization.
Lower values indicate less overlap across islands.
Best program similarity compares the current best programs across islands.
Population proximity measures similarity between island populations.
Exact-code overlap measures shared normalized programs; and Descendant proximity excludes the initial programs before measuring population
similarity.
Detailed definitions are provided in Appendix~\ref{app:similarity}.
}
\label{tab:initialization_redundancy}
\end{table}

Max--min initialization substantially reduces redundancy across all four
metrics.
In particular, island-best similarity decreases from $0.183$ to $0.062$,
while cross-island population proximity decreases from $0.424$ to
$0.086$.
The descendant-only population proximity, which excludes the initial
programs themselves, also decreases from $0.122$ to $0.085$.
This suggests that the effect of diverse initialization persists beyond
the initial seeds and leads to less overlapping island trajectories in
subsequent mutations.
Detailed definitions of the similarity metrics are provided in
Appendix~\ref{app:similarity}.

\paragraph{Effect on final performance.}
We next examine whether the greater cross-island diversity observed above translates into better final search performance. We compare three initialization strategies: using the same initial program for all islands, randomly selecting four programs from 20 generated candidates, and selecting four using our max--min criterion. For a fair comparison, all strategies use a total budget of 100 program generations, excluding auxiliary meta-prompt calls. Same-start initialization uses all 100 rounds for evolutionary search, while random and max--min initialization allocate 20 generations to initialization and the remaining 80 rounds to evolutionary search. To isolate the effect of initialization, all other components, including island selection and migration, are kept fixed.

\begin{table}[!htb]
\centering
\small
\setlength{\tabcolsep}{6pt}
\renewcommand{\arraystretch}{1.10}
\begin{tabular}{lr}
\toprule
Initialization & Mean $\pm$ SD $\uparrow$ \\
\midrule
Same & 0.700 $\pm$ 0.026 \\
Random & 0.682 $\pm$ 0.020 \\
Ours (Max--min) & \textbf{0.714 $\pm$ 0.022} \\
\bottomrule
\end{tabular}
\caption{Comparison of initialization strategies on Signal Processing
using Qwen3.8-27B over five random seeds.
Results report the mean $\pm$ sample standard deviation
of the final best score.}
\label{tab:initialization_ablation}
\end{table}

As shown in Table~\ref{tab:initialization_ablation}, our max--min
initialization achieves the highest mean final score, improving from
$0.700$ with the same initialization and $0.682$ with random
initialization to $0.714$.
Together with the redundancy analysis above, these results indicate that
diverse initialization can reduce overlapping search while maintaining
better final search performance.

\subsection{Trajectory Adaptive Island Selection}
\label{sec:selection_analysis}

\paragraph{Selection behavior and performance.}
We examine whether adapting computation allocation to the evolving states of different islands improves final search performance.
We compare round-robin, uniform random selection, UCB, and our trajectory adaptive island selection using the same four islands.
To isolate the effect of island selection, all policies use the same diverse initial programs, while migration is disabled.

Table~\ref{tab:selection_policy} reports both final search performance and the standard deviation of island selection counts. Our method achieves the highest final score, $0.659 \pm 0.044$, while also showing the largest variation in selection counts. Across the four policies, larger variation in selection counts is associated with better final performance in this experiment. This suggests that differentiating computation across islands according to their evolving states can make better use of a fixed search budget.

\begin{table}[!htb]
\centering
\small
\setlength{\tabcolsep}{6pt}
\renewcommand{\arraystretch}{1.10}
\begin{tabular}{lcc}
\toprule
Selection policy & Final score $\uparrow$ & Selection-count SD \\
\midrule
Round-robin & 0.628 $\pm$ 0.031 & 0.00 \\
Uniform     & 0.645 $\pm$ 0.036 & 5.49 \\
UCB         & 0.633 $\pm$ 0.036 & 3.48 \\
Ours        & \textbf{0.659 $\pm$ 0.044} & 17.33 \\
\bottomrule
\end{tabular}
\caption{
Comparison of island-selection policies using Qwen3.8-27B.
Final score reports mean $\pm$ sample standard deviation. 
Selection-count SD is the standard deviation of the selection counts across the four islands within each run, averaged across runs.
}
\label{tab:selection_policy}
\end{table}

\subsection{Growth Adaptive Migration}
\label{sec:migration_analysis}

\begin{figure*}[!t]
\centering
\begin{minipage}[t]{0.47\textwidth}
    \centering
    \includegraphics[width=\linewidth]{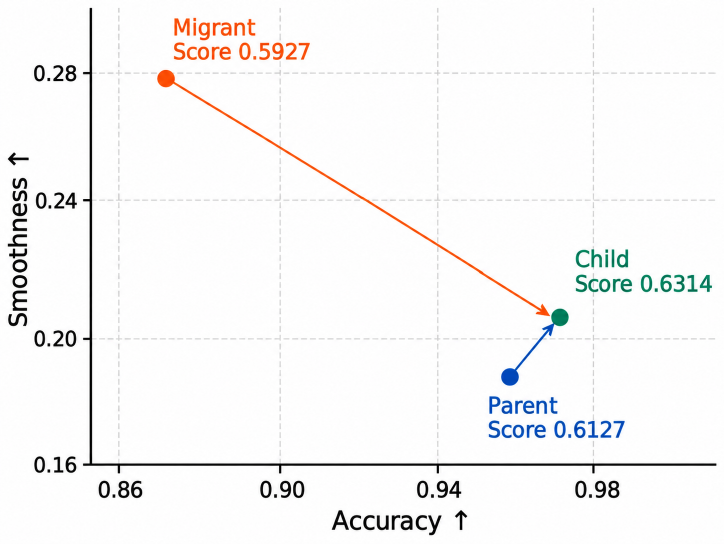}
    \small (a) Accuracy--smoothness trade-off
\end{minipage}
\hfill
\begin{minipage}[t]{0.47\textwidth}
    \centering
    \includegraphics[width=\linewidth]{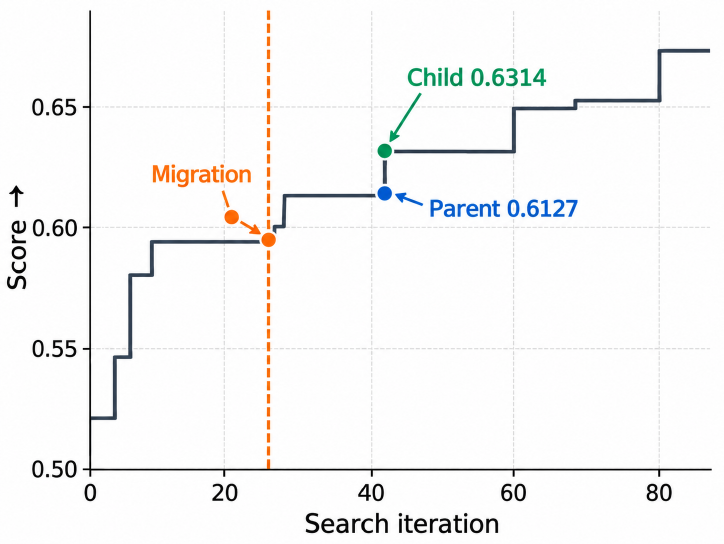}
    \small (b) Best-so-far score
\end{minipage}

\caption{
Qualitative example of migration on Signal Processing.
(a) The recipient parent and migrant exhibit complementary
strengths, and the resulting child improves both accuracy
and smoothness over the parent.
(b) Migration occurs at iteration 25, followed by
the discovery of an improved child at iteration 41.
}
\label{fig:migration_case}
\end{figure*}

We illustrate how migration enables islands to benefit
from complementary ideas using a case study on Signal
Processing. As shown in Figure~\ref{fig:migration_case}(a),
the recipient parent achieves higher accuracy, while
the migrated program provides greater smoothness,
despite having a lower overall score.

Following migration at iteration 25, a subsequent
LLM mutation produces a child at iteration 41 that
combines the strengths of both programs.
The child improves both accuracy and smoothness
over the parent, increasing the best score from
$0.6127$ to $0.6314$, as shown in
Figure~\ref{fig:migration_case}(b).

This example illustrates how migration can facilitate
the exchange of complementary ideas across islands,
allowing lower-scoring programs to contribute to
subsequent improvements.

\subsection{Component Ablation}
\label{sec:ablation}

We examine how the components of \ours contribute to final search
performance. We compare the full framework with three variants:
replacing diverse initialization with the same initial program,
replacing trajectory adaptive island selection with round-robin
and disabling migration, and combining both modifications.
All configurations are evaluated on Signal Processing over
five random seeds, with the remaining search settings fixed.

\begin{table}[!htb]
\centering
\footnotesize
\setlength{\tabcolsep}{3pt}
\renewcommand{\arraystretch}{1.15}
\begin{tabular}{@{}cccc@{}}
\toprule
\shortstack{Diverse\\init.} &
\shortstack{Adaptive\\selection} &
\shortstack{Adaptive\\migration} &
Final score $\uparrow$ \\
\midrule
 & & & 0.698 $\pm$ 0.032 \\
\checkmark & & & 0.703 $\pm$ 0.047 \\
 & \checkmark & \checkmark & 0.700 $\pm$ 0.026 \\
\checkmark & \checkmark & \checkmark &
\textbf{0.714 $\pm$ 0.022} \\
\bottomrule
\end{tabular}
\caption{
Component ablation on Signal Processing over five random seeds.
Checkmarks indicate the use of diverse initialization,
trajectory adaptive island selection, and growth adaptive migration.
Without these components, we use the same initial program,
round-robin selection, and no migration, respectively.
Results report mean $\pm$ sample SD.
}
\label{tab:ablation}
\end{table}

As shown in Table~\ref{tab:ablation}, the full framework achieves
the highest mean final score of $0.714$, outperforming all three
variants. Replacing diverse initialization reduces the mean score
to $0.700$, while replacing adaptive selection and migration
reduces it to $0.703$. Applying both modifications further reduces
the mean score to $0.698$. These results suggest that diverse
initialization and adaptive search decisions provide complementary
benefits when combined in \ours.

\section{Conclusion}
\label{sec:conclusion}

We introduced \ours, an LLM-guided evolutionary search framework that
focuses on how multiple island trajectories are initialized, developed,
and connected throughout the search.
Our method combines diverse initialization, trajectory adaptive island
selection, and growth adaptive migration to maintain distinct search
trajectories while allocating computation and exchanging information
according to their observed progress.
Experiments across multiple program optimization tasks show that our
method achieves competitive overall performance, while controlled
analyses demonstrate the benefits of diverse initialization and
trajectory-dependent computation allocation.
Together, these results highlight the importance of not only maintaining
multiple islands, but also actively managing how their search
trajectories evolve under a fixed search budget.
\section*{Limitations}

Our approach has several limitations.
First, recent growth is estimated from observed improvements and therefore
provides only a noisy proxy for the future potential of an island,
especially when improvements occur sparsely or abruptly.
Second, our initialization and migrant selection use embedding-based code
distance, which does not necessarily correspond to semantic or
algorithmic diversity.
Finally, our experiments cover a limited set of LLMs, program
optimization tasks, and search budgets.
Further evaluation is needed to determine how the proposed mechanisms
generalize to different models, larger numbers of islands, and
substantially longer search horizons.

\FloatBarrier

\bibliography{custom}

\appendix
\clearpage

\appendix

\lstdefinestyle{pythonstyle3}{
  language={},
  basicstyle=\ttfamily\fontsize{9}{11}\selectfont,
  columns=fullflexible,
  keepspaces=true,
  showstringspaces=false,
  showspaces=false,
  showtabs=false,
  tabsize=4,
  breaklines=true,
  breakatwhitespace=false,
  numbers=none,
  frame=single,
  framerule=0.4pt,
  framesep=6pt,
  rulecolor=\color{black!35},
  backgroundcolor=\color{black!2},
  xleftmargin=6.4pt,
  xrightmargin=6.4pt,
  aboveskip=0pt,
  belowskip=0pt,
  upquote=true
}
\lstdefinestyle{pythonstyle4}{
  style=pythonstyle3,
  language=Python,
  keywordstyle=\color{blue!55!black}\bfseries,
  commentstyle=\color{green!35!black},
  stringstyle=\color{orange!55!black},
  backgroundcolor=\color{blue!2}
}

This appendix provides implementation details of \ours,
experimental settings, and the prompt templates used in our framework:
\begin{itemize}[leftmargin=0.3cm]
    \item \textbf{Appendix~\ref{app:implementation}:}
    Implementation details of \ours and definitions of diversity metrics
    \item \textbf{Appendix~\ref{app:main_comparison_analysis}:}
    Additional analyses of Table~\ref{tab:main_gpt5}
    \item \textbf{Appendix~\ref{app:experimental_setup}:}
    Benchmark descriptions, evaluation metrics, and experimental configurations
    \item \textbf{Appendix~\ref{app:prompts}:}
    Prompt templates for initial program generation and LLM mutation
    \item \textbf{Appendix~\ref{app:signal_seed_examples}:}
    Signal Processing examples of selected initial programs
\end{itemize}

\section{Implementation Details}
\label{app:implementation}

\subsection{Diverse Initialization}
\label{app:initialization}

The distance in Equation~\ref{eq:diverse_initialization} is computed
from code embeddings $e(p)$ and $e(q)$ as
\begin{equation}
 d(p,q)=1-\frac{e(p)^\top e(q)}{\|e(p)\|_2\|e(q)\|_2}.
\end{equation}
We solve the max--min problem exactly by enumerating all subsets of
$\mathcal V$ of size $K=4$. Eligibility requires successful parsing,
execution, and the task-specific interface and feasibility checks;
fitness is not an additional ranking term in the diversity objective.
The selected source code and evaluations are cached for reuse across
comparison conditions. Each island's stored population has capacity 20.

Joint-generation experiments additionally remove candidates with
identical abstract syntax trees and embed the complete eligible
programs with Qwen3-Embedding-8B. Initialization stops if fewer than
$K$ valid, distinct programs remain. Generation templates appear in
Appendix~\ref{app:initial_prompt}; the historical pool and embedding
cache used in the initialization comparison are specified in
Appendix~\ref{app:initialization_protocol}.

\subsection{Trajectory Adaptive Island Selection}
\label{app:selection}

\paragraph{Normalization and initialization of statistics.}
To obtain $\widetilde{\mu}_i(t)$ and $\widetilde{g}_i(t)$ in
Section~\ref{sec:selection}, we apply min--max scaling separately to
$\hat{\mu}_i(t)$ and $\hat{g}_i(t)$ across islands. If all islands
have the same value for a statistic, its normalized values are set to
zero. The growth update uses $\rho=0.9$ and $\epsilon=10^{-8}$.
Before a finite previous best is available, the first finite
evaluation uses $\delta_{I_t}(t)=1$. The adaptive policies first
sample uniformly among islands with fewer than three recorded visits,
including the initial observation, before applying the selection index.

\subsection{Growth Adaptive Migration}
\label{app:migration}

\paragraph{Timing and transfer limits.}
The experimental implementation supplements the adaptive mechanism
in Section~\ref{sec:migration} with ring exchange every 20 mutation
rounds. Using the paper's island indices, the ring follows
$1\to2\to3\to4\to1$, with at most three programs per edge.
Adaptive migration uses $\tau=\sqrt{0.5}$ in the main-text growth
criterion, since the implementation compares squared growth values
with a ratio of 0.5. It becomes eligible after 20 iterations and at least five local evaluations in the recipient.
One qualifying check is sufficient, subject to a five-round interval
since that recipient's latest migration attempt, including ring
exchange. Each adaptive event transfers at most five programs.

\paragraph{Program selection and population update.}
In addition to excluding donor bests as specified in the main text,
we exclude code already present in the recipient. Migration uses
Qwen3-Embedding-8B. For geometry and Signal Processing, it embeds
code reachable from the task entry point when extraction is supported,
falling back to the full source otherwise; the systems tasks use
the full source. Historical initialization embeddings are cached
separately.

Transfers copy programs into the recipient and retain the donor's
copies. Low-fitness recipient programs are replaced while protecting
its best program and protected archive representatives. The actual
transfer count can be smaller than the cap when there are too few
eligible candidates or replaceable slots. All transfers in one ring
event are planned from the pre-event populations, so an incoming
program cannot be forwarded again within that event.

The growth update in Section~\ref{sec:migration} uses the number of
programs actually admitted and the pre-event growth estimates.
We synchronize the recipient's best-score baseline with the updated
population without awarding a local improvement reward. When meta
guidance is active, its global-best parent override takes precedence
over the migration follow-up parent choice
(Appendix~\ref{app:mutation_prompt}).

\subsection{Diversity Metrics}
\label{app:similarity}

\begin{figure*}[!ht]
    \centering
    \includegraphics[width=0.85\textwidth]{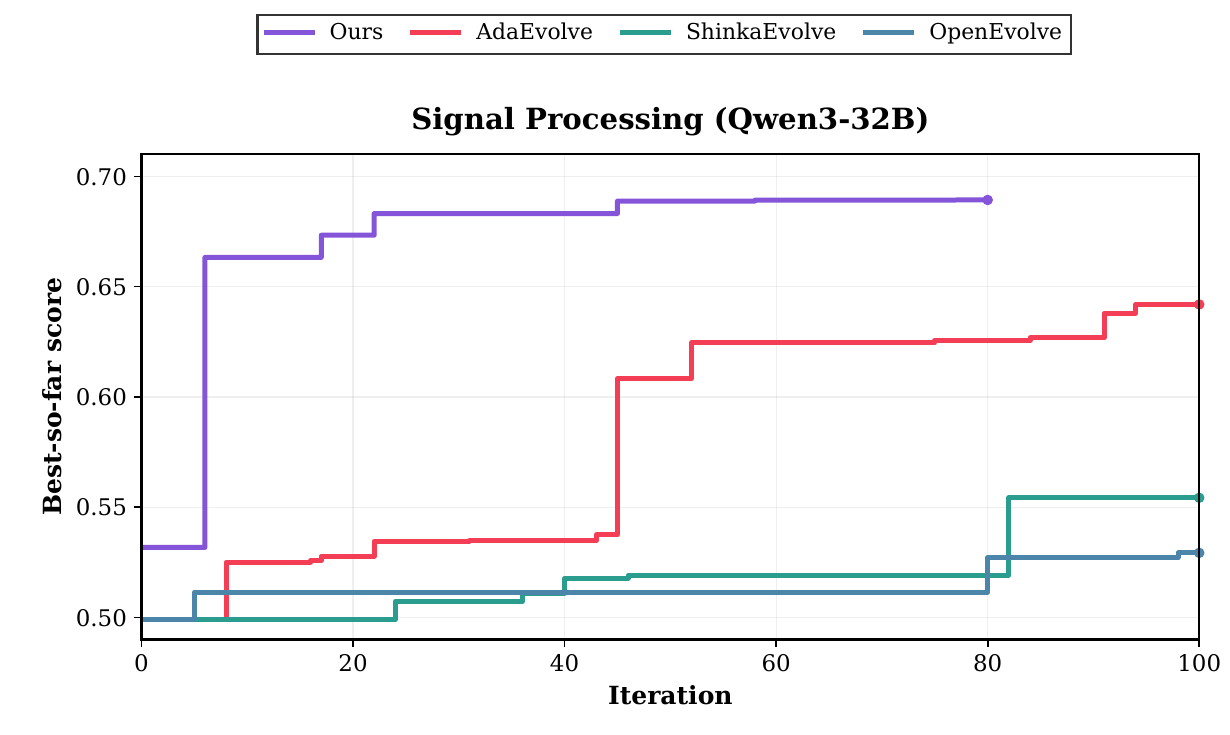}
    \caption{
    Example best-so-far score trajectories on Signal Processing
    using Qwen3-32B.
    Ours uses 80 evolutionary rounds after 20 initial
    program generations, while the baselines use
    100 evolutionary rounds.
    }
    \label{fig:signal_processing_compare}
\end{figure*}

The redundancy analysis uses lexical code similarity, independently
of the embedding distance used to select initial programs or migrants.
We extract each program's mutable \texttt{EVOLVE-BLOCK}, tokenize it
as Python, remove comments and layout tokens, and replace string and
numeric literals with type placeholders. Identifiers and operators
are retained. Let $S(p)$ be the set of consecutive five-token shingles
in this normalized sequence. We define
\begin{equation}
 \operatorname{sim}(p,q)=
 \frac{|S(p)\cap S(q)|}{|S(p)\cup S(q)|}.
\end{equation}
All four reported redundancy metrics lie in $[0,1]$, with smaller
values indicating less overlap.

\paragraph{Best program similarity.}
For each island, select the stored program with the highest evaluator
fitness. Average $\operatorname{sim}$ over the six unordered pairs of
island bests.

\paragraph{Population proximity.}
For $\mathcal D_{i,t}$ and $\mathcal D_{j,t}$, compare each program
with its most similar program in the other population. Average these
nearest-neighbor similarities over the programs in both populations,
giving each program equal weight. The reported proximity is the mean
over the six island pairs.

\paragraph{Exact code overlap.}
Let $H_i(t)$ be the set of complete normalized token sequences of
programs in $\mathcal D_{i,t}$. We average
$|H_i(t)\cap H_j(t)|/|H_i(t)\cup H_j(t)|$ over island pairs.
Equality is exact only after the normalization above; this metric
does not require byte-identical source files.

\paragraph{Descendant proximity.}
We recompute population proximity after excluding all iteration-zero
initial programs. This measures overlap among subsequently generated
programs without directly counting the shared starting code.

Run-level metrics are averaged equally across the cohort specified in
Appendix~\ref{app:initialization_protocol}. Missing round-20 checkpoints
were reconstructed from the preceding checkpoint and saved generation
records, with population sizes and best scores checked against the
iteration logs. These metrics quantify lexical overlap, not semantic
independence.

\section{Additional Analysis of the Main Comparison}
\label{app:main_comparison_analysis}

We provide additional examples from Signal Processing
and Circle Packing to illustrate the search behavior
and final solutions underlying the results in
Table~\ref{tab:main_gpt5}.

\paragraph{Signal Processing.}

Figure~\ref{fig:signal_processing_compare} illustrates
the evolution of the best-so-far score for each method
using Qwen3-32B.
In this example, \ours achieves a substantial improvement
early in the search and continues to improve with
additional generations, maintaining a higher best-so-far
score than the baselines.
The baselines exhibit different improvement patterns,
with some achieving substantial gains only in later
rounds.
These trajectories illustrate how the methods differ
in their search behavior beyond the final scores
reported in Table~\ref{tab:main_gpt5}.

\begin{figure*}[!ht]
    \centering
    \includegraphics[width=\textwidth]{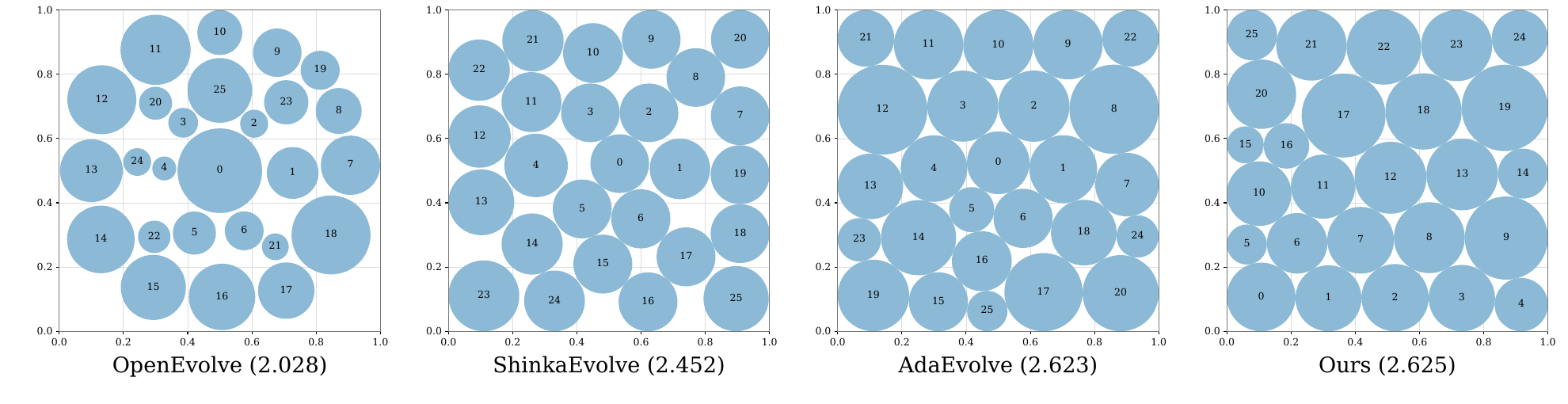}
    \caption{
    Example solutions for Circle Packing with $n=26$
    using Qwen3-32B.
    Each panel shows a circle arrangement in the unit
    square, with its corresponding score in parentheses.
    Higher scores indicate better solutions.
    }
    \label{fig:circle_packing_compare}
\end{figure*}
\paragraph{Circle Packing.}
Figure~\ref{fig:circle_packing_compare} visualizes
final solutions obtained by different methods on
Circle Packing with $n=26$ using Qwen3-32B.
Each program seeks to maximize the sum of circle radii
while maintaining a valid packing inside the unit square.
In the displayed examples, \ours achieves a score of
$2.625$, compared with $2.623$ for AdaEvolve,
$2.452$ for ShinkaEvolve, and $2.028$ for OpenEvolve.
These examples illustrate the geometric configurations
produced by the different methods, complementing
the aggregate performance comparison in
Table~\ref{tab:main_gpt5}.

\twocolumn[{
\section{Experimental Setup}
\label{app:experimental_setup}

\subsection{Benchmarks and Evaluation Metrics}
\label{app:benchmarks}

\begin{minipage}{\textwidth}
Table~\ref{tab:appendix_benchmarks} summarizes the tasks. Below, we
specify how valid outputs are scored and how reported objectives
relate to search fitness.

\medskip
\centering
\small
\setlength{\tabcolsep}{5pt}
\renewcommand{\arraystretch}{1.18}
\begin{tabular}{@{}p{0.18\textwidth}p{0.31\textwidth}p{0.39\textwidth}c@{}}
\toprule
Task & Evaluation setting & Reported metric & Goal \\
\midrule
\raggedright Heilbronn Triangle
  & 11 points in an equilateral triangle; all 165 point triples
  & Minimum triangle area divided by the containing triangle's area
  & $\uparrow$ \\
\raggedright Circle Packing (Square)
  & 26 non-overlapping circles in the unit square
  & Sum of radii of a valid packing
  & $\uparrow$ \\
\raggedright Signal Processing
  & Five noisy signals, 500--900 samples; window size 20
  & Weighted score combining filtering quality, smoothness,
    correlation, noise reduction, and execution success
  & $\uparrow$ \\
\raggedright PRISM
  & 50 placement cases; 5--9 GPUs with 80\,GB per GPU
  & Inverse mean worst-GPU KV-cache pressure plus successful-case fraction
  & $\uparrow$ \\
\raggedright EPLB
  & 11 next-window load evaluations; 288 replicas per layer on 32 GPUs
  & Mean of expert balancedness and a normalized inverse-time score
  & $\uparrow$ \\
\raggedright CloudCast
  & Five intra- and inter-provider broadcast scenarios
  & Total simulated transfer and instance cost across all scenarios
  & $\downarrow$ \\
\bottomrule
\end{tabular}
\captionsetup{hypcap=false}
\captionof{table}{Benchmark overview. Arrows indicate the preferred
direction of the reported metric. Geometry objectives and CloudCast
cost retain the units used in the main comparison.}
\label{tab:appendix_benchmarks}
\end{minipage}
\vspace{1.5em}
}]

\paragraph{Heilbronn Triangle.}
A program places 11 points inside a unit-side equilateral triangle.
After checking containment, the evaluator finds the smallest area
among all 165 point triples and divides it by the containing area,
$\sqrt{3}/4$. We report this area ratio; search additionally divides
it by the reference value $0.03652988988$.

\paragraph{Circle Packing (Square).}
The evaluator verifies that all 26 circles lie inside the unit square
and do not overlap, with numerical tolerance $10^{-6}$. The reported
objective is the sum of their nonnegative radii, recomputed from the
output. Invalid packings score zero. Search divides the sum by 2.635;
the main table reports the unscaled value.

\paragraph{Signal Processing.}
The five signal families are a sinusoid with trend, mixed frequencies,
a chirp, steps, and a random walk with trend, with Gaussian noise
standard deviations from 0.2 to 0.6. Programs receive the full noisy
array and window size 20; evaluation aligns the output at offset 19
and also uses the clean signal. This interface permits noncausal
filters.

For each successful case, $S$ counts output directional changes,
$L_{\mathrm{recent}}$ and $L_{\mathrm{avg}}$ are the final and mean
aligned absolute deviations from the noisy input, and $R$ counts
output reversals without a matching clean-signal reversal. The
composite quality is
\begin{align}
 J={}&0.3\min(S/50,2)+0.2\min(L_{\mathrm{recent}},2)\notag\\
     &+0.2\min(L_{\mathrm{avg}},2)+0.3\min(R/25,2),\notag\\
 c={}&(1+J)^{-1}.
\end{align}
The reported fitness is
\begin{align}
 F={}&0.4\bar c+\frac{0.2}{1+\bar S/20}
       +0.2\max(0,\bar r)\notag\\
     &+0.1\bar\eta+0.1u.
 \label{eq:appendix_signal_score}
\end{align}
Here $r$ is Pearson correlation with the clean signal, $\eta$ is
the fractional reduction in noise variance, clipped below at zero,
and $u$ is the successful-case fraction. The five terms reward
composite quality, smoothness, correlation, noise reduction, and
reliability, respectively. In the current evaluator, $\bar c$ averages
over all five cases with failures scored zero; the remaining averages
use successful cases. No successful case, or mean correlation below 0.1,
gives zero fitness. Execution time is not part of this score.

\paragraph{PRISM.}
Programs assign serving models to GPUs while respecting memory
limits and placing each model exactly once. For each GPU, KV-cache
pressure divides the sum of request-rate/latency-target ratios by
its memory remaining after model placement. The worst GPU determines
the case pressure; a GPU with no remaining memory receives a large
penalty. The reported score is $1/\bar v+u$, where $\bar v$ is mean
pressure over successful cases and $u$ is their fraction among all
50 cases. This measures placement quality rather than serving
throughput.

\paragraph{Expert Parallel Load Balancing (EPLB).}
A recorded trace contains 58 layers with 256 logical experts each.
For 11 consecutive transitions, a program uses a 100-step workload
window to assign 288 physical replicas per layer to 32 GPUs; the
mapping is evaluated on the next window. Replica counts and coverage
of loaded experts are checked. For physical-expert loads $L_{\ell,e}$,
balancedness is $B=\sum_\ell\operatorname{mean}_e L_{\ell,e}/
\sum_\ell\max_e L_{\ell,e}$. The score is
$\tfrac12(\bar B+0.002/\bar t)$, where $\bar t$ is mean rebalancing
plus simulated-inference time in seconds and both means cover the
11 transitions. Higher scores reward balanced loads and shorter
runtime; GPU-level balancedness is a separate diagnostic.

\paragraph{CloudCast.}
Programs construct broadcast routes for three intra-provider and two
inter-provider scenarios involving AWS, Azure, and GCP. Every
destination must receive every partition along valid network paths.
The simulator accounts for bandwidth, two virtual machines per
region, transfer charges, and instance charges. We report total cost
$C_{\mathrm{total}}$ across the five cases; search maximizes
$1/(1+C_{\mathrm{total}})$. Invalid broadcasts receive zero fitness.

\subsection{Experimental Configuration}
\label{app:configuration}
\label{app:experiment_provenance}

\paragraph{Generation and evaluation.}
Our GPT-5 runs use medium reasoning effort. The Signal Processing
analyses use Qwen3.8-27B through vLLM on RTX A6000 GPUs, with
thinking disabled, temperature 0.7, a 16,384-token output limit,
and a 32,768-token context. They use four islands. 
Program evaluations use three repeats with median
aggregation and a 360-second outer timeout. Final evaluations reuse
the benchmark suite. Reported mean $\pm$ SD summarizes search
seeds using sample SD; evaluation repeats are not additional seeds.
Retries, and meta-generation calls are outside the mutation-round count.

\paragraph{Initialization comparison.}
\label{app:initialization_protocol}
Same-start copies the original program to all islands.
Random samples four eligible programs after 20 candidates generation.
Max--min uses the fixed diversity-selected set from generated 20 candidates. 

Table~\ref{tab:initialization_ablation} uses five seeds per condition:
Same reports final evaluations at round 100, while Random and Max--min
use 20 first generations and 80 rounds.
Table~\ref{tab:initialization_redundancy} uses the same five
seeds and Qwen3.8-27B at round 20 of Same--start, and at round 40 of Random and Max--min. 
Similarity definitions are given in Appendix~\ref{app:similarity}.

\paragraph{Island-selection comparison.}
\label{app:allocation}
Table~\ref{tab:selection_policy} retains ten seeds per policy, each
with 100 rounds, the same four diverse initial programs, and both
migration and meta generation disabled. 
Round-robin is cyclic.
Uniform samples islands independently. 
UCB follows the implementation of \citet{cemri2026adaevolve}. 
Our policy uses $C=0.3$.
The allocation statistic is the population SD of the four islands'
mutation attempt counts, averaged over runs. It measures unequal
allocation and differs from the sample SD of score across seeds.

\paragraph{Component variants.}
\label{app:component_ablation}
Each uses five seeds, a total budget of 100 program generations. Same
copies the original program; Diverse uses the shared four-program
cache. Adaptive migration includes both native ring and growth-triggered
migration with growth mixing (Appendix~\ref{app:migration}); No
migration disables both.

\section{Prompt Templates}
\label{app:prompts}

\subsection{Initial Program Generation}
\label{app:initial_prompt}

The templates below summarize the actual prompt structure using
braced placeholders for task-specific content. Evaluator and source
code placeholders are expanded to the corresponding code in a
request. Technique guidance is written in advance; the programs and
their proposed approaches are generated by the LLM.

\paragraph{Independent proposals.}
Each request starts from the normal mutation prompt for the reference
program, with no earlier attempts or meta-generated idea, and appends
the assignment in Figure~\ref{fig:appendix_independent_prompt}.

\begin{figure*}[!tp]
\centering
\begin{lstlisting}[style=pythonstyle3]
# Diverse Initial Program Generation

## Assigned Strategy
You are producing one member of a diverse initial population.
Your assigned algorithmic family is: {algorithmic_family}

## Material Diversity Requirement
Implement the assigned family in executable code. An unchanged program, or
changes only to comments, names, formatting, or constants, do not constitute
a different approach. Do not retain the original algorithm as the active path.

## Execution Contract
Preserve {function_signature}, {output_contract}, and {domain_constraints}.
The program must run independently and respect {execution_limit}.

## Output
Return a substantive SEARCH/REPLACE edit to {reference_program}.
Every SEARCH block must match the reference source exactly, including whitespace.
\end{lstlisting}

\caption{Independent initialization assignment, appended to the
reference program's mutation prompt. Braced fields are replaced with
task-specific instructions.}
\label{fig:appendix_independent_prompt}
\end{figure*}

Strategy assignments include explicit constructions, deterministic
sampling, local or global search, feasibility-preserving
parameterizations, and hybrid construction and refinement. Signal
Processing assignments instead emphasize suitable filter families,
such as robust median filters, adaptive exponential smoothers,
state-space estimators, local polynomial filters, and multiscale
filters. These assignments guide generation; execution and validity
are checked separately.

\paragraph{Joint proposals.}
The joint-generation variant replaces the individual assignment
with a request for a complementary set
(Figure~\ref{fig:appendix_joint_prompt}).

\begin{figure*}[!tp]
\centering
\begin{lstlisting}[style=pythonstyle3]
# Joint Initial Population Design

## System Message
You are an expert in {task_domain}. Jointly design an initial population of
independent Python programs for an island-based program evolution system.
Optimize both solution quality and complementarity. Return executable programs.

## Problem and Interface
{problem_description}
{objective_and_constraints}
{function_signature_and_output_contract}

## Authoritative Evaluator
```python
\end{lstlisting}

\begin{lstlisting}[style=pythonstyle4]
{evaluator_code}
\end{lstlisting}

\begin{lstlisting}[style=pythonstyle3]
```

## Initial Reference Program
```python
\end{lstlisting}

\begin{lstlisting}[style=pythonstyle4]
{reference_program}
\end{lstlisting}

\begin{lstlisting}[style=pythonstyle3]
```

## Technique Guidance
{task_specific_techniques_and_feasibility_pitfalls}
These are optional suggestions, not mandatory algorithm assignments.
Programs may share an optimizer if their formulation or parameterization differs.

## Joint Population Design
Generate exactly four independent programs as a complementary set.
Different constants, random seeds, names, or comments alone do not count as
algorithmic diversity. Each program should pursue a strong valid solution.
Do not deliberately weaken a program to make it different.

## Execution Contract
Preserve the required interface. Include all imports and complete source code.
Respect {execution_limit} and use only {available_dependencies}.
Validate the result before returning it. Do not modify or bypass the evaluator.

## Output Format
Return one JSON object with key "programs", containing exactly four objects.
Each object has "island_id", "approach", "distinctive_feature", and "code".
Each code string must be complete and independently executable.
Do not use ellipses, placeholders, or references to another program.
\end{lstlisting}

\caption{Joint initialization template. Each request produces four
candidate programs; evaluation and max--min selection subsequently
determine the final initial population.}
\label{fig:appendix_joint_prompt}
\end{figure*}

For example, geometry guidance covers constrained optimization,
compact geometric parameterizations, multistart search, and exact
constraint checking. In Heilbronn Triangle, it explicitly requires
checking all 165 point triples. Responses are checked for complete
JSON, all four code strings, and Python syntax before evaluation;
a partial program is not treated as a valid starting program.

\subsection{LLM Mutation}
\label{app:mutation_prompt}

\paragraph{Parent and references.}
After island selection, the database samples a parent and up to four
reference programs. Normal mutation uses local references in these
experiments. Parent sampling balances high-fitness programs with
novelty-based exploration; its intensity depends on the island's
growth statistic. Migration follow-up can prioritize received
programs in the reference slots. The builder inserts code, metrics,
and the sampling-mode instruction into the structure in
Figure~\ref{fig:appendix_mutation_prompt}.

\begin{figure*}[!tp]
\centering
\begin{lstlisting}[style=pythonstyle3,escapeinside={(*@}{@*)}]
# LLM Mutation

## System Message
{task_specific_expertise_objective_and_constraints}

## Current Solution Information
- Main metrics: {evaluator_metrics}
- Focus areas: {improvement_areas}

## Program Generation History
{previous_attempts_if_available}
{reference_programs_with_scores}

## Current Solution
```python
\end{lstlisting}

\begin{lstlisting}[style=pythonstyle4]
{parent_program}
\end{lstlisting}

\begin{lstlisting}[style=pythonstyle3,escapeinside={(*@}{@*)}]
```
{parent_metrics_and_sampling_mode}

## Search Guidance
{evaluator_feedback_if_available}
{active_meta_generated_idea_if_available}
{sibling_outcome_summary_if_available}
{retry_error_information_if_available}

## Task
Improve the primary evaluator score while preserving the interface and validity
constraints. Explain the proposed changes and respect {execution_limit}.
Every SEARCH block must match the current solution character for character,
including all whitespace and indentation. Use the exact edit format below.

## Output Format
(*@\texttt{\detokenize{<<<<<<< SEARCH}}@*)
{exact_source_fragment}
(*@\texttt{\detokenize{=======}}@*)
{replacement_source_fragment}
(*@\texttt{\detokenize{>>>>>>> REPLACE}}@*)
\end{lstlisting}

\caption{Mutation prompt template, including conditional search
guidance and the required SEARCH/REPLACE edit format.}
\label{fig:appendix_mutation_prompt}
\end{figure*}

The parser applies edits to the parent and evaluates the resulting
complete program. Feedback is conditional: a request does not contain
every earlier program or a full search transcript. The sibling summary
reports outcomes of previous mutations of the same parent, including
score changes, rather than inserting all sibling code. Retry requests
can include parsing or execution errors and can resample the parent.

\paragraph{Meta-generated guidance.}
When enabled, a stagnation tracker can request a separate LLM review.
That request contains the task, evaluator, current global-best code
and score, and previously tried ideas with feedback. A fixed review
template asks the model to analyze the objective, metrics,
constraints, input--output structure, and improvement opportunities
before proposing alternative approaches. The LLM returns structured
ideas with fields for the idea, implementation description,
optimization target, cautions, and approach type. The fixed analysis
framework and technique suggestions are supplied by the implementation;
the concrete review and proposed ideas are LLM-generated.

An active idea is inserted into subsequent mutation guidance, and the
controller uses the global-best program as the parent for that guided
mutation. This mechanism is enabled in the full-method and factor
ablation configurations, but disabled in the isolated island-selection
comparison. It is triggered by search state rather than invoked on
every mutation, so enabled runs can make different numbers of
auxiliary review calls.

\FloatBarrier

\raggedbottom
\section{Signal Processing Initialization Examples}
\label{app:signal_seed_examples}

The four examples below are the actual filtering implementations in
the cached diverse initialization used by the Signal Processing
component studies. They were selected from the historical candidate
pool described in Appendix~\ref{app:initialization_protocol}, before
evolution with Qwen3.8-27B. They are initial programs, rather than
high-scoring descendants or outputs of the later joint-generation
protocol.

The selected set covers four approaches:
\begin{itemize}[leftmargin=0.3cm]
    \item \textbf{Local polynomial regression:} independently fits a
    polynomial to each window and evaluates it at the right endpoint.
    \item \textbf{Adaptive Kalman filtering:} recursively estimates
    level and velocity, adapting measurement noise from residuals.
    \item \textbf{Median plus EMA:} applies a rolling median for
    outlier robustness, followed by exponential smoothing.
    \item \textbf{Bounded-trend Holt filtering:} recursively updates
    level and trend while limiting changes in the trend estimate.
\end{itemize}
These distinctions concern the executable algorithms, not changes
to names or constants. Selection uses embedding distance; the
algorithm-family descriptions are an interpretation of the selected
code, not additional labels supplied to the selection procedure.

Figures~\ref{fig:appendix_seed_window_filters}--\ref{fig:appendix_seed_holt}
reproduce the filtering functions and the required Kalman class.
Long comments and docstrings are omitted and lines are wrapped for
presentation; executable statements and default parameters are
preserved. The common benchmark runner, synthetic-data helper, and
diagnostic printing are omitted. In the original runner,
\texttt{run\_signal\_processing} passes the noisy input and window
size through \texttt{process\_signal} to the corresponding filter,
then returns its output under \texttt{filtered\_signal}. All examples
use NumPy. The benchmark uses a window size of 20, so an input of
length $n$ produces $n-19$ filtered samples.

\begin{figure*}[!tp]
\centering
\begin{lstlisting}[style=pythonstyle4,basicstyle=\ttfamily\fontsize{9}{10}\selectfont]
import numpy as np

# (a) Local polynomial regression
def causal_polynomial_filter(x, window_size=20, polynomial_degree=1):
    if len(x) < window_size:
        raise ValueError(
            f"Input signal length ({len(x)}) must be >= "
            f"window_size ({window_size})")
    if polynomial_degree < 0 or polynomial_degree >= window_size:
        raise ValueError(
            f"Polynomial degree ({polynomial_degree}) must be non-negative "
            f"and less than window_size ({window_size})")
    output_length = len(x) - window_size + 1
    y = np.zeros(output_length)
    t = np.arange(window_size)
    evaluation_point = window_size - 1
    for i in range(output_length):
        window = x[i:i + window_size]
        coefficients = np.polyfit(t, window, polynomial_degree)
        y[i] = np.polyval(coefficients, evaluation_point)
    return y

# (b) Rolling median followed by exponential smoothing
def causal_median_ema_filter(x, window_size=20, ema_alpha=0.2):
    n = len(x)
    if n == 0:
        return np.array([])
    if window_size < 1:
        raise ValueError("window_size must be at least 1")
    if n < window_size:
        raise ValueError(
            f"Input signal length ({n}) must be >= window_size ({window_size})")
    median_window_size = window_size
    if median_window_size % 2 == 0:
        median_window_size += 1
    if median_window_size > n:
        median_window_size = n if n % 2 != 0 else n - 1
        if median_window_size < 1:
            median_window_size = 1
    half_window = (median_window_size - 1) // 2
    median_filtered = np.zeros(n - window_size + 1)
    for i in range(len(median_filtered)):
        start_idx = max(0, i + window_size - median_window_size)
        end_idx = i + window_size
        current_median_window = x[start_idx:end_idx]
        if len(current_median_window) < median_window_size:
            median_filtered[i] = np.median(x[:end_idx])
        else:
            median_filtered[i] = np.median(current_median_window)
    filtered_signal = np.zeros_like(median_filtered)
    if len(median_filtered) > 0:
        filtered_signal[0] = median_filtered[0]
        for i in range(1, len(median_filtered)):
            filtered_signal[i] = (
                ema_alpha * median_filtered[i]
                + (1 - ema_alpha) * filtered_signal[i - 1])
    return filtered_signal
\end{lstlisting}

\caption{Two selected initial filters with different window-based
operations: endpoint polynomial fitting and robust median filtering
with recursive smoothing. Code is shown before any evolutionary
mutation.}
\label{fig:appendix_seed_window_filters}
\end{figure*}

\begin{figure*}[!tp]
\centering
\begin{lstlisting}[style=pythonstyle4,basicstyle=\ttfamily\fontsize{9}{10}\selectfont]
import numpy as np

class KalmanFilter:
    def __init__(
        self, initial_pos, initial_vel=0.0, process_noise_pos=0.01,
        process_noise_vel=0.001, measurement_noise=0.1, adaptive_alpha=0.05
    ):
        self.x = np.array([[initial_pos], [initial_vel]])
        self.F = np.array([[1, 1], [0, 1]])
        self.Q = np.array([[process_noise_pos, 0], [0, process_noise_vel]])
        self.H = np.array([[1, 0]])
        self.R = np.array([[measurement_noise]])
        self.initial_R = measurement_noise
        self.adaptive_alpha = adaptive_alpha
        self.P = np.eye(2) * 1.0

    def predict(self):
        self.x = self.F @ self.x
        self.P = self.F @ self.P @ self.F.T + self.Q

    def update(self, z):
        z_array = np.array([[z]])
        y = z_array - self.H @ self.x
        S = self.H @ self.P @ self.H.T + self.R
        K = self.P @ self.H.T @ np.linalg.inv(S)
        self.x = self.x + K @ y
        self.P = (np.eye(self.x.shape[0]) - K @ self.H) @ self.P
        residual_variance = y.T @ np.linalg.inv(S) @ y
        self.R = (
            self.adaptive_alpha * residual_variance
            + (1 - self.adaptive_alpha) * self.R)
        self.R = np.clip(self.R, self.initial_R / 10, self.initial_R * 10)

def kalman_filter_adaptive(x, window_size=20):
    if len(x) < window_size:
        raise ValueError(
            f"Input signal length ({len(x)}) must be >= "
            f"window_size ({window_size})")
    initial_pos = np.mean(x[:window_size])
    if window_size > 1:
        initial_vel = (x[window_size - 1] - x[0]) / (window_size - 1)
    else:
        initial_vel = 0.0
    kf = KalmanFilter(
        initial_pos=initial_pos, initial_vel=initial_vel,
        process_noise_pos=0.01, process_noise_vel=0.001,
        measurement_noise=0.5, adaptive_alpha=0.02)
    output_start_idx = window_size - 1
    y = np.zeros(len(x) - output_start_idx)
    for i, measurement in enumerate(x):
        kf.predict()
        kf.update(measurement)
        if i >= output_start_idx:
            y[i - output_start_idx] = kf.x[0, 0]
    return y
\end{lstlisting}

\caption{Selected adaptive Kalman initialization, including its
state-estimation class. The program carries level and velocity
estimates across observations and adjusts measurement noise using
the innovation residual.}
\label{fig:appendix_seed_kalman}
\end{figure*}

\begin{figure*}[!tp]
\centering
\begin{lstlisting}[style=pythonstyle4,basicstyle=\ttfamily\fontsize{9}{10}\selectfont]
import numpy as np

def holt_double_exponential_filter(
    x, window_size=20, alpha=0.5, beta=0.1, trend_bound_factor=0.5
):
    if len(x) < window_size:
        raise ValueError(
            f"Input signal length ({len(x)}) must be >= "
            f"window_size ({window_size})")
    if not (0 < alpha <= 1 and 0 < beta <= 1):
        raise ValueError(
            "Alpha and Beta must be between 0 and 1 "
            "(exclusive of 0, inclusive of 1).")
    if trend_bound_factor <= 0:
        raise ValueError("Trend bound factor must be positive.")
    output_length = len(x) - window_size + 1
    y = np.zeros(output_length)
    initial_window = x[0:window_size]
    L = np.mean(initial_window)
    if window_size > 1:
        t_indices = np.arange(window_size)
        A = np.vstack([t_indices, np.ones(window_size)]).T
        m, c = np.linalg.lstsq(A, initial_window, rcond=None)[0]
        T = m
    else:
        T = 0.0
    for i in range(output_length):
        current_window_end_value = x[i + window_size - 1]
        L_prev = L
        T_prev = T
        L = alpha * current_window_end_value + (1 - alpha) * (L_prev + T_prev)
        T_candidate = beta * (L - L_prev) + (1 - beta) * T_prev
        if T_prev != 0:
            max_trend_change = abs(T_prev * trend_bound_factor)
            T_change = T_candidate - T_prev
            if T_change > max_trend_change:
                T_candidate = T_prev + max_trend_change
            elif T_change < -max_trend_change:
                T_candidate = T_prev - max_trend_change
        T = T_candidate
        y[i] = L
    return y
\end{lstlisting}

\caption{Selected Holt initialization. The filter maintains level and
trend estimates, limiting each nonzero trend's change to a fraction
of its previous magnitude. The returned value is the updated level.}
\label{fig:appendix_seed_holt}
\end{figure*}

\FloatBarrier

\end{document}